%% file: main.tex
\def\year{2021}\relax
\documentclass[letterpaper]{article} 
\usepackage{aaai21}  
\usepackage{times}  
\usepackage{helvet} 
\usepackage{courier}  
\usepackage[hyphens]{url}  
\usepackage{graphicx} 
\usepackage{natbib}  
\usepackage{caption} 
\usepackage{adjustbox}
\usepackage{multirow}

\usepackage{microtype}

\usepackage{xspace}
\usepackage{mathtools}
\usepackage{amssymb}
\usepackage{rotating}
\usepackage{footnote}
\usepackage{color, colortbl}
\usepackage{xstring}
\usepackage{comment}

\usepackage{float}
\restylefloat{table}
\usepackage{array,booktabs,makecell}
\usepackage{graphicx}
\usepackage{appendix}
\usepackage[normalem]{ulem}
\usepackage{tcolorbox}
\usepackage{enumitem}

\usepackage{scalerel,xparse}
\usepackage{siunitx}

\newcolumntype{H}{>{\setbox0=\hbox\bgroup}c<{\egroup}@{}}

\usepackage{arydshln}

\PassOptionsToPackage{hyphens}{url}\usepackage{hyperref}

\begin{document}

\title{Decay No More:\\  A Persistent Twitter Dataset for Learning Social Meaning}

\author{Chiyu Zhang  ~~~~Muhammad Abdul-Mageed ~~~ El Moatez Billah Nagoudi \\ 
 \textnormal{Deep Learning \& Natural Language Processing Group \\The University of British Columbia} \\
  \tt chiyuzh@mail.ubc.ca, \tt \{muhammad.mageed,moatez.nagoudi\}@ubc.ca}

\maketitle
\begin{abstract}
With the proliferation of social media, many studies resort to social media to construct datasets for developing social meaning understanding systems. For the popular case of Twitter, most researchers distribute tweet IDs without the actual text contents due to the data distribution policy of the platform. One issue is that the posts become increasingly inaccessible over time, which leads to unfair comparisons and a temporal bias in social media research. To alleviate this challenge of data decay, we leverage a paraphrase model to propose a new \textit{persistent} English Twitter dataset for social meaning (PTSM). PTSM consists of $17$ social meaning datasets in $10$ categories of tasks. We experiment with two SOTA pre-trained language models and show that our PTSM can substitute the actual tweets with paraphrases with \textit{marginal} performance loss.\footnote{Our data is available at: \url{https://github.com/chiyuzhang94/PTSM}.} 
\end{abstract}

\input{intro}
\input{related_work}
\input{experiment}

\input{results}
\input{conclusion}

\bibliography{literature}

\end{document}

%% file: intro.tex
\section{Introduction}
\textit{Social meaning} is substantiated in socio-pragmatics, and refers to intended meaning in real-world communication~\cite{thomas2014meaning} and how utterances should be interpreted within the social context in which they are produced. Aspects of social meaning include emotion recognition~\cite{mohammad-2018-semeval}, irony detection~\cite{van-hee2018semeval}, sarcasm detection~\cite{riloff2013sarcasm, bamman2015contextualized}, hate speech identification~\cite{waseem-2016-hateful}, and stance identification~\cite{mohammad-2016-semeval}. A successful social meaning comprehension system can ameliorate a wide range of NLP applications. For example, a dialogue system with knowledge of social meaning can provide more engaging reactions when it realizes human emotions. 

\begin{figure}[ht]
\begin{centering}
 \includegraphics[width=0.85\linewidth]{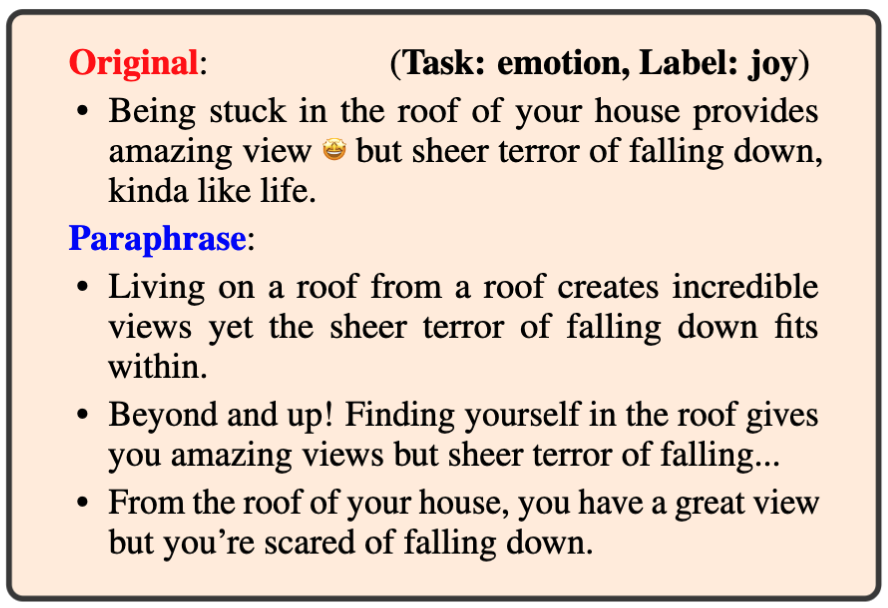}
  \caption{Paraphrase example from Emo\textsubscript{Moham}.}
  \label{fig:para}
\end{centering}
\end{figure}

With the proliferation of social media, billions of users share content in various forms. Social media platforms, such as Twitter, allow users to express their opinions, discuss topics, and connect with friends, among other practices~\cite{farzindar2015natural}. As a result, social media offers abundant resources for social meaning understanding. Over the years, researchers have developed a number of labelled datasets to train (semi-)supervised machine learning models. According to the data redistribution policy of a social media platform such as Twitter, a post's actual content cannot be shared with third parties. Hence, many studies only distribute the IDs of posts (e.g., tweets). The challenge with this set up, however, substantial social media posts become inaccessible over time due to deletion, protection, etc.~\cite{zubiaga2018longitudinal, assenmacher2021benchmarking}. We empirically characterize this issue of data inaccessibility by attempting to re-collect the tweet contents of six social meaning datasets. As Table~\ref{tab:crawl} shows, we could only acquire, on average, $73\%$ of the tweets. This data decay leads to a serious issue concerning the lack of replicability in social media studies. The issue of data inaccessibility also introduce a temporal bias (as older data sets become significantly unavailable) and leads to unfair comparisons. In other words, it is difficult to compare models trained on differently sized datasets. In this work, we propose a \textit{persistent} Twitter dataset for social meaning (PTSM) to alleviate these issues. We use a state-of-the-art (SOTA) Transformer-based pre-trained model, i.e., T5~\cite{raffel2020exploring}, as backbone to train a paraphrasing model with diverse parallel data. We then use the trained model to produce sentential paraphrases of training samples. We then benchmark our PTSM with two SOTA pre-trained language models (PLM), RoBERTa~\cite{liu2019roberta} and BERTweet~\cite{nguyen-etal-2020-bertweet}. Our experiment results show that we can replace the actual tweets with their paraphrases without sacrificing performance: On average, our paraphrase-based models are only $\sim 1.70$ $F_1$ below models trained with original data across the $17$ datasets.

To summarize, we make the following \textbf{contributions}:
\textbf{(1)} We introduce PTSM, a persistent Twitter dataset for social meaning comprising $17$ different datasets, whose accessibility is enhanced via paraphrasing. To the best of our knowledge, this is the first work to employ SOTA paraphrase methods to tackle occasional data decay for social media research. \textbf{(2)} We develop a Transformer-based model for paraphrasing social media data. \textbf{(3)} We benchmark our PTSM with two SOTA PLMs and demonstrate the promise of training tweet classifiers with paraphrases only.

\input{tables/data_inaccess}


%% file: tables/data_inaccess.tex
\begin{table}[h]
\centering
\tiny
\begin{tabular}{@{}lcrrc@{}}
\toprule
\multicolumn{1}{c}{\textbf{}}                     & \textbf{Prod. year} & \multicolumn{1}{c}{\textbf{Orig.}} & \multicolumn{1}{c}{\textbf{Retr.}} & \multicolumn{1}{c}{\textbf{Decay \%}} \\ \midrule
\citet{riloff2013sarcasm}        & 2013                  & $3.0$K                                 & $1.8$K                                  & 0.41                                 \\
\citet{ptavcek2014sarcasm}       & 2014                  & $100.0$K                               & $89.3$K                                 & 0.11                                 \\
\citet{rajadesingan2015sarcasm}  & 2015                  & $91.0$K                                & $51.6$K                                 & 0.43                                 \\
\citet{bamman2015contextualized} & 2015                  & $19.5$K                                & $14.8$K                                 & 0.34                                 \\
\citet{waseem-2016-hateful}        & 2016                  & $16.9$K                                & $10.9$K                                 & 0.36                                 \\
\citet{rosenthal-2017-semeval}        & 2017                  & $50.3$K                                & $48.2$K                                 & 0.04                                 \\ \bottomrule
\end{tabular}
\caption{Issue of data inaccessibility. These six datasets were distributed by their authors via tweet IDs. \textbf{Orig.:} original size of each dataset. \textbf{Retr.:} data we successfully collected via Twitter API in November, 2020. \textbf{Decay \%:} percentage of inaccessible tweets. }\label{tab:crawl}
\end{table}

%% file: related_work.tex
\vspace{-5pt}
\section{Related Work}
Paraphrasing aims at rewriting or rephrasing a text while maintaining its original semantics. Most previous works introduce paraphrasing as a way to augment training data and hence alleviate data sparsity in machine learning models. To reduce the high degree of lexical variation,~\citet{petrovic2012using, li2016using} produce paraphrases of tweets by replacing words of original text with their synonyms based on WordNet and word embedding vector closeness.~\citet{beddiar2021data} introduce a back-translation method to augment training data for hate speech detection. Different to the previous works, we utilize paraphrases to develop a tweet classifier without any subsequent use of the original tweets while training our downstream models. Our main objective is to tackle the data decay issue in machine learning of social media. For that, we offer a unified paraphrase dataset for training social meaning models that is directly comparable to original training data for a host of tasks. Our work has affinity to work aiming at facilitating meaningful model comparisons such as the general language understanding evaluation (GLUE) benchmark~\cite{wang2019glue} and SuperGLUE~\cite{sarlin2020superglue}, but we focus on availing training data that enable model building in the first place.~\citet{barbieri-2020-tweeteval} introduce TweetEval, a benchmark for tweet classification evaluation, but they are not able to share more than $50$K tweets per dataset due to Twitter distribution policies. Different to them, we are able to provide unlimited numbers of paraphrases for future research. 

%% file: experiment.tex
\section{Persistent Dataset for Social Meaning}
\input{gold_dataset}


\input{t5_para}

\subsection{Generating PTSM}
We apply the trained paraphrasing model on the Train split of each of our $17$ social meaning datasets, using top-$p$ sampling~\cite{holtzman2019curious} with $p=0.95$ to generate $10$ paraphrases for each gold sample. We then select paraphrases of a given tweet based on the following criteria: \textbf{(i)} To remove any `paraphrases' that are just copies or near-duplicates of the original tweet, we use a simple tri-gram similarity method where we exclude any paraphrases whose similarity with the original tweet is $> 0.95$. \textbf{(ii)} To ensure that paraphrases are not totally different from the original tweets, we remove any generations whose tri-gram similarity with the original tweet $=0$. \textbf{(iii)} We then sort the remaining paraphrases by their tri-gram similarities with the original tweet and descendingly add paraphrase into set $P$. To populate $P$, we proceed as follows: we loop over paraphrases one by one, calculating the similarity of each each other paraphrase in $P$. If the paraphrase is $> 0.50$ similar to any item in $P$, we do not it add into $P$.  Ultimately, each original tweet associates with a set of paraphrases $P$. This process results in a paraphrasing dataset \texttt{Para-Train-Clean}. We present example paraphrases in Figure~\ref{fig:para} and Table~\ref{tab:para_exam_app}. We observe that the paraphrase model fails to generate emojis since emojis are out-of-vocabulary for the original T5 model, but the model can preserve the overall semantics of the original tweet.
To explore effect of the size of paraphrase data on the downstream tasks, we extract $1$, $2$, $4$, and $5$ paraphrases from \texttt{Para-Train-Clean} for each Train gold sample in each of our $17$ tasks. We refer to the resulting datasets as \texttt{Para$1$}, \texttt{Para$2$}, \texttt{Para$4$}, and \texttt{Para$5$}. Although the main puprpose of this paper is to solve the data decay issue of training data, we also provide a paraphrase dataset for Dev and Test splits of each of our $17$ social meaning datasets. For each Dev or Test gold sample, we extract one paraphrase sample. We refer to the resulting datasets as \texttt{Para-Dev} and \texttt{Para-Test}. Table~\ref{tab:para_n_data} shows the distribution of the resulting paraphrase datasets.\footnote{Note that some samples may not have enough paraphrases to construct Para$2$, Para$4$, and Para$5$.} 
 \input{tables/para_data}
\input{tables/para_sample_app}

%% file: gold_dataset.tex
\subsection{Social Meaning Datasets}\label{sec:smpb}
\input{tables/dataset}
We collect $17$ datasets representing $10$ different social meaning tasks, as follows:~\footnote{To facilitate reference, we give each dataset a name.}

\noindent\textbf{Crisis awareness.}~We use \texttt{Crisis\textsubscript{Oltea}} ~\cite{olteanu2014crisislex}, a corpus for identifying whether a tweet is related to a given disaster or not. 
    
 \noindent     \textbf{Emotion.} We utilize \texttt{Emo\textsubscript{Moham}}, introduced by~\citet{mohammad-2018-semeval}, for emotion recognition. We use the version adapted in \citet{barbieri-2020-tweeteval}.
 
\noindent\textbf{Hateful and offensive language.} We use \texttt{Hate\textsubscript{Bas}}~\cite{basile-2019-semeval}, \texttt{Hate\textsubscript{Waseem}}~\cite{waseem-2016-hateful}, \texttt{Hate\textsubscript{David}}~\cite{davidson-2017-hateoffensive}, and \texttt{Offense\textsubscript{Zamp}}~\cite{zampieri-2019-predicting}.

\noindent     \textbf{Humor.} We use the humor detection datasets \texttt{Humor\textsubscript{Potash}}~\cite{potash-2017-semeval} and \texttt{Humor\textsubscript{Meaney}}~\cite{meaney2021hahackathon}.

\noindent\textbf{Irony.} We utilize \texttt{Irony\textsubscript{Hee-A}} (irony detection) and  \texttt{Irony\textsubscript{Hee-B}} (irony type identification) from \citet{van-hee2018semeval}. 
  
\noindent     \textbf{Sarcasm.} 
We use four sarcasm datasets from \texttt{Sarc\textsubscript{Riloff}}~\cite{riloff2013sarcasm}, \texttt{Sarc\textsubscript{Ptacek}}~\cite{ptavcek2014sarcasm}, \texttt{Sarc\textsubscript{Rajad}}~\cite{rajadesingan2015sarcasm}, and \texttt{Sarc\textsubscript{Bam}}~\cite{bamman2015contextualized}. 
 
\noindent\textbf{Sentiment.} We employ the three-way sentiment analysis dataset from \texttt{Senti\textsubscript{Rosen}}~\cite{rosenthal-2017-semeval} and a binary sentiment analysis dataset from \texttt{Senti\textsubscript{Thel}}~\cite{thelwall2012sentiment}. 

\noindent\textbf{Stance.} We use \texttt{Stance\textsubscript{Moham}}, a stance detection dataset from \citet{mohammad-2016-semeval}. The task is to identify the position of a given tweet towards a target of interest. 


\textbf{Data Crawling and Preparation.} We use the Twitter API~\footnote{\url{https://developer.twitter.com/}} to crawl datasets which are available only in tweet ID form. 
Before we paraphrase the data or use it in our various experiments, we normalize each tweet by replacing the user names and hyperlinks to the special tokens `USER' and `URL', respectively. This ensures our paraphrased dataset will never have any actual usernames or hyperlinks, thereby protecting user identity. For datasets collected based on hashtags by original authors, we also remove the seed hashtags from the original tweets. For datasets originally used in cross-validation, we acquire $80\%$ Train, $10\%$ Dev, and $10\%$ Test via random splits. For datasets that had training and test splits but not development data, we split off $10\%$ from training data into Dev. The splits of each dataset are presented in Table~\ref{tab:gold_data}.

%% file: tables/dataset.tex
\begin{table}[!ht]
\centering
\tiny
\begin{tabular}{@{}llrrrr@{}}
\toprule
\multicolumn{1}{c}{\textbf{Task}}                & \multicolumn{1}{l}{\textbf{Classes}}                            & \multicolumn{1}{c}{\textbf{Train}} & \multicolumn{1}{c}{\textbf{Dev}} & \multicolumn{1}{c}{\textbf{Test}} & \multicolumn{1}{c}{\textbf{Total}} \\ \midrule
Crisis\textsubscript{Oltea}
& \{on-topic, off-topic,\}                                        & $48.0$K                             & $6.0$K                            & $6.0$K                             & $60.0$K                             \\
Emo\textsubscript{Moham}
& \{anger, joy, opt., sad.\}                               & $3.3$K                              & $374$                              & $1.4$K                             & $5.0$K                              \\
Hate\textsubscript{Bas}
& \{hateful, none\}                                        & $9.0$K                              & $1.0$                            & $3.0$                             & $13.0$K                             \\
Hate\textsubscript{Waseem}
& \{racism, sexism, none\}                                        & $8.7$K                              & $1.1$K                            & $1.1$K                             & $10.9$K                             \\
Hate\textsubscript{David}
& \{hate, offensive, neither\}                                    & $19.8$K                             & $2.5$K                            & $2.5$K                            & $24.8$K                            \\
Humor\textsubscript{Potash}
& \{humor, not humor\}                          & $11.3$K                             & $660$                              & $749$                               & $12.7$K                             \\
Humor\textsubscript{Meaney}
&                     \{humor, not humor\}                                             & $8.0$K                             & $1.0$K                              & $1.0$K                             & $10.0$K                             \\
Irony\textsubscript{Hee-A}
& \{ironic, not ironic\}                                          & $3.5$K                             & $384$                              & $784$                               & $4.6$K                             \\
Irony\textsubscript{Hee-B}
& \{IC, SI, OI, NI\} & $3.5$K                              & $384$                              & $784$                               & $4.6$K                              \\
Offense\textsubscript{Zamp}
& \{offensive, not offensive\}                                   & $11.9$K                             & $1.3$K                           & $860$                               & $14.1$K                            \\
Sarc\textsubscript{Riloff}
& \{sarcastic, non-sarcastic\}                   & $1.4$K                             & $177$                              & $177$                               &$1.8$K                             \\
Sarc\textsubscript{Ptacek}
&    \{sarcastic, non-sarcastic\}                                                              & $71.4$K                             & $8.9$K                            & $8.9$K                             & $89.3$K                          \\
Sarc\textsubscript{Rajad}
&   \{sarcastic, non-sarcastic\}                                                               & $41.3$K                            & $5.2$K                          & $5.2$K                           & $51.6$K                          \\
Sarc\textsubscript{Bam}
&          \{sarcastic, non-sarcastic\}                                                        & $11.9$K                             & $1.5$K                          & $1.5$K                             & $14.8$K                             \\
Senti\textsubscript{Rosen}
& \{neg., neu., pos.\}                                 & $42.8$K                            & $4.8$K                           & $12.3$K                         & $59.8$K                           \\
Senti\textsubscript{Thel}
& \{neg., pos.\}                                 & $900$                            & $100$                           & $1.1$K                         & $2.1$K                           \\
Stance\textsubscript{Moham}
& \{against, favor, none\}                                        & $2.6$K                              & $292$                              & $1.3$K                             & $4.2$K                             \\ \bottomrule
\end{tabular}%
\caption{Gold social meaning datasets. \textbf{opt.:}: Optimism, \textbf{sad.:} Sadness, \textbf{IC:} Ironic by clash, \textbf{SI:} Situational irony, \textbf{OI:} Other irony, \textbf{NI:} Non-ironic, \textbf{neg.:} Negative, \textbf{Neu.:} Neutral, \textbf{pos.:} Positive.}\label{tab:gold_data}
\end{table}

%% file: t5_para.tex
\subsection{Praphrase Model}\label{sec:para} 

In order to train our paraphrasing model, we collect four paraphrase datasets from PIT-2015~\cite{xu2015semeval}, LanguageNet~\cite{lan2017continuously}, Opusparcus~\cite{creutz2018open}, and Quora Question Pairs (QQP)~\cite{iyer2017first}. We only keep sentence pairs with a high semantic similarity as follows: (1) For PIT-2015, we extract sentence pairs with semantic similarity labels of $5$ and $4$. (2) For LanguageNet, we keep sentence pairs obtained with similarity labels of $4$, $5$, and $6$. (3) For Opusparcus, we take pairs whose similarity labels are $4$. (4) For QQP, we extract sentence pairs with the `duplicate' label. Table~\ref{tab:para_train_data} presents the data size of each corpus after filtering. We then merge all extracted samples (a total of $625,097$ pairs) and split them into Train, Dev, and Test ($80\%$, $10\%$, and $10\%$). 
   \input{tables/para_train_data}

For the modelling, we fine-tune a pre-trained T5\textsubscript{Base}~\cite{raffel2020exploring} on the Train split for $20$ epochs with a constant learning rate of $3e-4$ and a maximal sequence length of $512$. We evaluated the model on the Dev set at the end of each epoch and identified the best model ($28.18$ BLEU score). 


%% file: tables/para_train_data.tex
\begin{table}[!h]
\renewcommand\thetable{3}
\centering
\small
\begin{tabular}{@{}lcrH@{}}
\toprule
\multicolumn{1}{c}{\textbf{Dataset}} & \textbf{Domain}      & \multicolumn{1}{c}{\textbf{\# of samples}} &  \textbf{Sequence l.}\\ \midrule
PIT-2015                             & Tweet                & 3,789         &                              \\
LanguageNet                          & Tweet                & 12,988        &                              \\
Opusparcus                           & Video subtitle       & 462,846       &                              \\
QQP                                  & Quora                & 149,263       &                              \\ \hdashline
Total                                & \multicolumn{1}{r}{} & 625,097                                     \\ \bottomrule
\end{tabular} 
\caption{Paraphrasing datasets.}\label{tab:para_train_data}
\end{table}

%% file: tables/para_data.tex
\begin{table}[h]
\renewcommand\thetable{4} 
\centering
\tiny
\begin{tabular}{@{}llrrr:rr@{}}
\toprule
\multicolumn{1}{c}{\textbf{Task}}            & \multicolumn{1}{c}{\textbf{Para1}} & \multicolumn{1}{c}{\textbf{Para2}} & \multicolumn{1}{c}{\textbf{Para4}} & \multicolumn{1}{c}{\textbf{Para5}} & \multicolumn{1}{c}{\textbf{ParaD}} & \multicolumn{1}{c}{\textbf{ParaT}} \\ \midrule
Crisis\textsubscript{Oltea} & $48.0$K                            & $86.8$K                            & $120.7$K                           & $123.9$K                           & $6.0$K                                & $6.0$K                                 \\
Emo\textsubscript{Moham}    & $3.3$K                             & $6.3$K                             & $10.9$K                            & $12.2$K                            & $374$                                 & $1.4$K                                 \\
Hate\textsubscript{Bas}     & $9.0$K                             & $17.6$K                            & $31.1$K                            & $35.1$K                            & $1.0$                                 & $3.0$                                  \\
Hate\textsubscript{Waseem}  & $8.7$K                             & $16.6$K                            & $28.3$K                            & $31.7$K                            & $1.1$K                                & $1.1$K                                 \\
Hate\textsubscript{David}   & $19.8$K                            & $38.2$K                            & $65.5$K                            & $73.4$K                            & $2.5$K                                & $2.5$K                                 \\
Humor\textsubscript{Potash} & $11.3$K                            & $21.8$K                            & $38.3$K                            & $44.0$K                            & $660$                                 & $749$                                  \\
Humor\textsubscript{Meaney} & $8.0$K                             & $15.7$K                            & $28.7$K                            & $33.0$K                            & $1.0$K                                & $1.0$K                                 \\
Irony\textsubscript{Hee-A}  & $3.5$K                             & $6.6$K                             & $11.4$K                            & $12.8$K                            & $384$                                 & $784$                                  \\
Irony\textsubscript{Hee-B}  & $3.5$K                             & $6.6$K                             & $11.5$K                            & $12.9$K                            & $384$                                 & $784$                                  \\
Offense\textsubscript{Zamp} & $11.9$K                            & $23.0$K                            & $39.6$K                            & $44.3$K                            & $1.3$K                                & $860$                                  \\
Sarc\textsubscript{Riloff}  & $1.4$K                             & $2.7$K                             & $4.6$K                             & $5.2$K                             & $177$                                 & $177$                                  \\
Sarc\textsubscript{Ptacek}  & $71.4$K                            & $138.9$K                           & $242.1$K                           & $272.4$K                           & $8.9$K                                & $8.9$K                                 \\
Sarc\textsubscript{Rajad}   & $41.3$K                            & $78.3$K                            & $131.5$K                           & $146.6$K                           & $5.2$K                                & $5.2$K                                 \\
Sarc\textsubscript{Bam}     & $11.9$K                            & $22.4$K                            & $37.5$K                            & $41.6$K                            & $1.5$K                                & $1.5$K                                 \\
Senti\textsubscript{Rosen}  & $42.8$K                            & $84.3$K                            & $154.8$K                           & $178.1$K                           & $4.8$K                                & $12.3$K                                \\
Senti\textsubscript{Thel}   & $900$                              & $1.7$K                             & $3.1$K                             & $3.5$K                             & $100$                                 & $1.1$K                                 \\
Stance\textsubscript{Moham} & $2.6$K                             & $4.7$K                             & $6.4$K                             & $6.6$K                             & $292$                                 & $1.3$K                                 \\ \bottomrule
\end{tabular}
\caption{Distribution of PTSM. \textbf{Para$n$} indicates that the Train set with varying paraphrase size. \textbf{ParaD:} Para-Dev, \textbf{ParaT:} Para-Test.}\label{tab:para_n_data}
\end{table}

%% file: tables/para_sample_app.tex
\begin{table*}[]
\renewcommand\thetable{5} 
\scriptsize
\centering
\begin{tabular}{llc}
\toprule
\textbf{Original Tweet}                                                                                                                                                & \textbf{Paraphrase}                                                                                   & \textbf{Label}                                 \\ \midrule
                                                                                                                                                                      & You guys are horrible, avoid MMT                                                                       & { }                        \\
                                                                                                                                                                      & what I am doing is in my control, \#AvoidMMT, you guys are terrifying                                  & { }                        \\ 
                                                                                                                                                                      & You guys are \#terrorist. I have used everything I have to do.                                         & { }                        \\
\multirow{-4}{*}{\begin{tabular}[c]{@{}l@{}}USER but what I am doing is in my control, \\ \#AvoidMMT , you guys are \#terrible\end{tabular}}                           & You guys are awful, but I am going to stop doing it.                                                   & \multirow{-4}{*}{{ anger}} \\\hline
                                                                                                                                                                      & The 46th wedding I've ruined. When I hit 50 I can retire. It's nice to see yo                          &                                                \\
                                                                                                                                                                      & Here's the 47th wedding I've ruined. If I'm old enough to go on the 40s I can get married.             &                                                \\
\multirow{-3}{*}{This is the 47th wedding I've ruined. When I hit 50 I can retire.}                                                                                    & After a single wedding, I drew 47 weddings, and before I hit 50 I can retire"                          & \multirow{-3}{*}{humor}                        \\\hline
                                                                                                                                                                      & Sorry to disturb you. I have absolutely no idea what time I'll be on cam tomorrow.                     &                                                \\
\multirow{-2}{*}{\begin{tabular}[c]{@{}l@{}}Sorry guys I have absolutely no idea \\ what time i'll be on cam tomorrow but will keep you posted.\end{tabular}}          & I have absolutely no idea what time I'll be on camera tomorrow but I'll keep you posted                & \multirow{-2}{*}{sadness}                      \\\hline
                                                                                                                                                                      & I'll buy you Dunkin' Donuts for \$5.                                                                  &                                                \\
                                                                                                                                                                      & Who wants to go with me for my tattoo tomorrow? I'll buy you a Dunkin' Donuts.                         &                                                \\
\multirow{-3}{*}{\begin{tabular}[c]{@{}l@{}}Who wants to go with me to get my tattoo tomorrow? \\ I'll buy you Dunkin doughnuts\end{tabular}}                          & Who wants to go with me to get my tattoo tomorrow?                                                     & \multirow{-3}{*}{neutral}                      \\\hline
                                                                                                                                                                      & The day before class please eat beans, onions and garlic. Also see the videos                          &                                                \\
                                                                                                                                                                      & The Day Before Class. You should make that meal, (do you think).                                      &                                                \\
\multirow{-3}{*}{\begin{tabular}[c]{@{}l@{}}USER May I suggest, that you have a meal that is made with \\ beans, onions \& garlic, the day before class.\end{tabular}} & If you can eat just the day before class, make a wonderful meal with garlic, onions and beans.         & \multirow{-3}{*}{joy} \\                        
        \toprule               
\end{tabular}%
\caption{More examples of paraphrases in PTSM.} \label{tab:para_exam_app}
\end{table*}

%% file: results.tex
\input{tables/para_res}
\section{Experiment and Result}\label{sec:result}

\subsection{Implementation}
We evaluate the quality of the $17$ paraphrased Train datasets in Table~\ref{tab:para_n_data} via fine-tuning two Transformer-based pre-trained language models, i.e, RoBERTa\textsubscript{Base}~\cite{liu2019roberta} and BERTweet\textsubscript{Base}~\cite{nguyen-etal-2020-bertweet}. We utilize the checkpoints released by Huggingface\footnote{\url{https://huggingface.co/models}}. 
 For Crisis\textsubscript{Oltea} and Stance\textsubscript{Moham}, we append the topic term behind the post content, separate them by an `[SEP]' token, and set maximal sequence length to $72$. For the rest of tasks, we set maximal sequence length to $64$. For all the tasks, we pass the hidden state of `[CLS]' token from the last Transformer encoder layer through two feed-forward layers (a linear layer with a $tanh$ activation function followed by another linear layer with a $softmax$ activation function) to predict the classification label. Cross-Entropy is used to calculate the training loss. Following~\citet{nguyen-etal-2020-bertweet}, we use Adam with a peak learning rate of $1e-5$ and a weight decay of $0.01$ to optimize the model and fine-tune each task for $20$ epochs with early stopping ($patience = 5$ epochs). The training batch size is $32$. We compare the model trained on PTSM to ones fine-tuned on the original gold Train set with the same hyper-parameters. We run three times with random seeds for all downstream fine-tuning, \textit{reporting the average of these three runs}. All downstream task models are fine-tuned on an Nvidia V$100$ GPUs ($32$G). 
 For individual task, we typically identify the best model on each respective Dev set and evaluate its performance on blind Test. We present the average Test macro-averaged $F_1$ over the three runs as mentioned, and introduce a global metric averaging the macro-$F_1$ scores over the $17$ datasets.

\subsection{Results}
We use our PTSM to investigate the viability of using paraphrased training data instead of gold training sets.  We fine-tune PLMs on the PTSM Train sets with varying paraphrase sizes (referred to as P$n$ in Table~\ref{tab:para_res}) but evaluate on the original Dev and Test sets for all of the individual tasks. As Table~\ref{tab:para_res} shows, although none of our paraphrase-based models exceed the corresponding baseline model that is fine-tuned on gold (i.e., original) Train sets in term of average $F_1$, the paraphrase-based models either slightly exceeds or approaches performance of the gold models on individual datasets. Regarding the effect of paraphrase data size, we find P2 Train to perform best both for RoBERTa and BERTweet models as compared to other amounts of paraphrase data. This shows that while doubling paraphrase data size is useful, more paraphrases do not help the models. RoBERTa-P2 (the best setting of paraphrase-based RoBERTa fine-tuning) obtains an average $F_1$ of $73.36$, which is $1.66$ less than RoBERTa-Gold (while outperforming the latter on Humor\textsubscript{Potash-17} and Senti\textsubscript{Rosen-17}). BERTweet-P2 (the best setting of paraphrase-based BERTweet fine-tuning) underperforms BERTweet-Gold with $1.87$ average $F_1$ (i.e., $74.93$). We also observe our BERTweet-P1 model achieves a sizable improvement of $3.36$ $F_1$ on Sarc\textsubscript{Riloff} over the gold model. These findings demonstrate that \textbf{(i)} we can replace social gold data (which can become increasingly \textit{inaccessible} over time) with paraphrase datasets (which are \textit{persistent}) \textbf{(ii)} without sacrificing much performance. In addition, we also fine-tune PLMs on the combination of gold Train and our paraphrase Train sets (referred to as P$n$+G in Table~\ref{tab:para_res}), but find average $F_1$ scores of our models to still remain below the models fine-tuned on gold data only. We hypothesize that one limitation of our paraphrased Train sets is the lack of emojis. However, analyzing presence of emoji contents for each of the datasets, we find that our paraphrase-based models are able to acquire comparable performance to models trained with original datasets that do not employ any emojis (e.g., none of the training samples of Senti\textsubscript{Rosen-17}, Senti\textsubscript{Thel}, and Stance\textsubscript{Moham} uses emojis). To further investigate the issue, we fine-tune PLMs on a version of the gold Train set after removing emojis. Here, we observe a slight degradation of performance: RoBERTa and BERTweet each obtains an average $F_1$ of $74.70$ and $76.49$, respectively (see Tabel~\ref{tab:no_emoji}).
These findings reflect the effect of emoji symbols on social meaning detection. This suggests we can enhance our paraphrase data by inserting the same emojis as original data. We cast this as future work. 
\input{tables/no_emoji}


Our experiments show that performance of a model trained on PTSM is on par with that of a model trained on gold data. Further, we investigate a scenario where we do not have access to any gold data for development nor test sets. In other words, we evaluate model performance on paraphrased Dev and Test sets under both gold and paraphrased training settings. We do so by fine-tuning RoBERTa on gold and P$n$ Train sets, using Para-Dev to identify the best model and testing on blind Para-Test. Regardless of the source of training data (gold or paraphrased), we find that models incur a significant performance drop as Table~\ref{tab:para_test} shows. That is, models evaluated on Para-Test drop $7.98$ $F_1$ points (gold-trained model) and $4.84$ $F_1$ points (best paraphrase-trained models) as compared to their respective counterparts evaluated on gold Test. We note that the scenario of no gold test data is not realistic, since it is usually fine to release gold test dataset (much less than $50$K data points in most cases). 


 
\input{tables/para_test}

%% file: tables/para_res.tex
\begin{table*}[ht]
\tiny
\centering
\begin{tabular}{lr:rrrr:rrrr|r:rrrr:rrrr}
\toprule
\multicolumn{1}{c}{\multirow{2}{*}{\textbf{Task}}}                                                                    & \multicolumn{9}{c}{\textbf{RoBERTa}}                                                                                                                                                                                                                                                                                                                                   & \multicolumn{9}{c}{\textbf{BERTweet}}                                                                                                                                                                                                                                                                                                                                                                  \\ \cmidrule(l){2-10} \cmidrule(l){11-19}
\multicolumn{1}{c}{}                                                                                         & \multicolumn{1}{c:}{\textbf{Gold}} & \multicolumn{1}{c}{\textbf{P1}}    & \multicolumn{1}{c}{\textbf{P2}}                                 & \multicolumn{1}{c}{\textbf{P4}}                                 & \multicolumn{1}{c:}{\textbf{P5}}    & \multicolumn{1}{c}{\textbf{P1+G}}  & \multicolumn{1}{c}{\textbf{P2+G}} & \multicolumn{1}{c}{\textbf{P4+G}} & \multicolumn{1}{c|}{\textbf{P5+G}} & \multicolumn{1}{c:}{\textbf{Gold}} & \multicolumn{1}{c}{\textbf{P1}}    & \multicolumn{1}{c}{\textbf{P2}}                                 & \multicolumn{1}{c}{\textbf{P4}}                                 & \multicolumn{1}{c:}{\textbf{P5}}                               & \multicolumn{1}{c}{\textbf{P1+G}}  & \multicolumn{1}{l}{\textbf{P2+G}} & \multicolumn{1}{c}{\textbf{P4+G}} & \multicolumn{1}{c}{\textbf{P5+G}} \\ \midrule
Crisis\textsubscript{O.}                                                                 & 95.96                          & \textbf{95.38} & 95.09                                                        & 95.18                                                        & 95.00                           & \textbf{95.88} & 95.85                             & 95.55                             & 95.17                             & 95.58                          & 94.90                                                        & 95.11                              & \textbf{95.26}                              & 95.17                                                        & \underline{\textbf{95.75}} & \underline{95.60}                             & 95.47                             & \underline{95.63}                             \\
Emo\textsubscript{M.}                                                                    & 77.61                          & 76.29                           & \textbf{77.16}                              & 77.02                                                        & 76.59                           & \underline{78.54} & \underline{77.65}                             & \underline{\textbf{78.82}}                    & \underline{78.54}                             & 80.37                          & 77.61                                                        & 78.06                              & 77.66                                                        & \textbf{78.62}                              & \textbf{80.71} & 80.29                             & 78.91                             & 79.56                             \\
Hate\textsubscript{B.}                                                                     & 49.74                          & 44.70                           & 46.48                                                        & 45.81                                                        & \textbf{47.89} & \textbf{48.78} & 47.99                             & 48.26                             & 48.21                             & 56.51                          & \textbf{54.01}                              & 52.41                              & 51.71                                                        & 52.04                                                        & \textbf{54.88} & 54.53                             & 53.01                             & 51.89                             \\
Hate\textsubscript{W.}                                                                  & 56.84                          & 54.22                           & \textbf{55.00}                              & 54.89                                                        & 54.92                           & 56.64 & \textbf{56.67}                    & 56.17                             & 55.46                             & 57.07                          & 55.33                                                        & 55.41                              & \underline{\textbf{59.33}} & 55.90                                                        & 56.42          & \textbf{56.87}                    & 56.10                             & 56.14                             \\
Hate\textsubscript{D.}                                                                   & 77.03                          & 75.14                           & 73.79                                                        & 74.43                                                        & \textbf{75.20} & \textbf{75.81} & 75.57                             & 74.66                             & 74.66                             & 77.57                          & 75.11                                                        & \textbf{75.75}    & 74.35                                                        & 75.08                                                        & \textbf{77.21} & 74.70                             & 76.00                             & 75.97                             \\
Humor\textsubscript{P.}                                                                 & 54.66                          & 52.92                           & \underline{\textbf{55.81}} & 54.18                                                        & 52.59                           & \underline{55.11}          & \underline{\textbf{56.10}}                    & 53.25                             & 50.64                             & 57.56                          & \textbf{52.77}                              & 52.65                              & 52.11                                                        & 50.92                                                        & \textbf{56.13} & 53.42                             & 53.77                             & 54.34                             \\
Humor\textsubscript{M.}                                                                 & 92.61                          & 90.96                           & \textbf{92.25}                              & 91.20                                                        & 91.96                           & 91.59          & 92.08                             & \textbf{92.45}                    & 91.57                             & 94.21                          & 93.31                                                        & 93.30                              & \textbf{93.76}                              & 92.67                                                        & 93.95          & 94.00                             & 93.57                             & \textbf{94.16}                    \\
Irony\textsubscript{H-A}                                                                  & 73.13                          & 69.96                           & 70.46                                                        & 70.52                                                        & \textbf{71.14} & 72.51 & \textbf{72.72}                    & 72.22                             & 72.06                             & 76.82                          & \textbf{76.26}                              & 75.42                              & 75.08                                                        & 75.37                                                        & \underline{\textbf{77.90}} & \underline{77.43}                             & \underline{77.04}                             & 75.60                             \\
Irony\textsubscript{H-B}                                                                  & 51.56                          & 46.23                           & \textbf{48.52}                              & 46.92                                                        & 45.31                           & \textbf{50.54} & 49.20                             & 49.79                             & 47.17                             & 56.84                          & 47.89                                                        & 51.18                              & 49.98                                                        & \textbf{51.78}                              & 56.24          & \textbf{56.45}                    & 50.72                             & 55.63                             \\
Offense\textsubscript{Z}                                                                & 80.67                          & 77.49                           & \textbf{80.22}                              & 80.18                                                        & 79.41                           & 80.39          & 80.20                             & \textbf{80.77}                    & 79.94                             & 79.74                          & 77.63                                                        & 78.89                              & 79.29                                                        & \underline{\textbf{80.15}} & 78.39          & 78.56                             & \textbf{79.47}                    & 79.32                             \\
Sarc\textsubscript{Ri.}                                                                  & 75.28                          & 71.04                           & 70.50                                                        & 71.22                                                        & \textbf{72.64} & 71.44          & 71.69                             & 73.39                             & \textbf{74.52}                    & 76.61                          & \underline{\textbf{79.97}} & \underline{78.61} & \underline{78.81}                           & 75.97                                                        & \textbf{\underline{80.19}} & \underline{77.01}                             & \underline{77.60}                             & \underline{77.21}                             \\
Sarc\textsubscript{P.}                                                                  & 95.59                          & 91.67                           & 93.48                                                        & 94.03                                                        & \textbf{94.35} & \textbf{95.34} & \textbf{95.34}                    & 95.11                             & 95.16                             & 96.74                          & 92.96                                                        & 94.34                              & 94.62                                                        & \textbf{94.70}                              & \textbf{96.34} & 96.04                             & 95.85                             & 95.71                             \\
Sarc\textsubscript{Ra.}                                                                   & 85.64                          & 80.82                           & \textbf{82.34}                              & 82.20                                                        & 82.29                           & \textbf{84.90} & 84.71                             & 84.00                             & 84.29                             & 86.97                          & 84.08                                                        & 84.99                              & 84.81                                                        & \textbf{85.31}                              & \textbf{86.95} & 86.55                             & 86.14                             & 86.19                             \\
Sarc\textsubscript{B.}                                                                     & 80.01                          & 77.33                           & \textbf{78.18}                              & 77.69                                                        & 77.09                           & \textbf{79.88} & 79.49                             & 78.91                             & 78.56                             & 82.59                          & 80.45                                                        & 80.60                              & \textbf{80.93}                              & 80.32                                                        & \underline{\textbf{82.92}} & 82.11                             & 81.99                             & 81.98                             \\
Senti\textsubscript{R.}                                                                  & 70.78                          & 70.59                           & \underline{\textbf{71.44}} & 70.26                                                        & 69.99                           & \underline{\textbf{71.19}} &  \underline{71.15}                             & 70.30                             & 70.76                             & 72.05                          & \textbf{71.23}                              & 70.21                              & 69.75                                                        & 69.98                                                        & 71.67          & 71.19                             & 71.17                             & \textbf{71.78}                    \\
Senti\textsubscript{T.}                                                                   & 88.99                          & 87.52                           & \textbf{88.12}                              & 87.83                                                        & 87.79                           & \underline{\textbf{89.04}} & 88.46                             & 88.48                             & 87.62                             & 89.27                          & 88.18                                                        & 89.00                              & 88.49                                                        & \underline{\textbf{89.28}} & 89.11          & \underline{89.35}                             & 89.16                             & \underline{\textbf{89.64}}                    \\
Stance\textsubscript{M.}                                                                 & 69.28                          & 67.81                           & 68.27                                                        & \underline{\textbf{69.76}} & 69.07                           & 68.56 & \textbf{68.64}                    & 68.44                             & 67.54                             & 69.11                          & 67.30                                                        & \textbf{67.92}    & 66.96                                                        & 66.76                                                        & \underline{69.12}          & \underline{\textbf{69.83}}                    & 68.11                             & 66.74                             \\
\cdashline{1-19} \multicolumn{1}{c}{\textbf{Average}} & 75.02                          & 72.36                           & \textbf{73.36}                              & 73.14                                                        & 73.13                           & \textbf{74.48} & 74.33                             & 74.15                             & 73.64                             & 76.80                          & 74.65                                                        & \textbf{74.93}    & 74.88                                                        & 74.71                                                        & \textbf{76.70} & 76.11                             & 75.53                             & 75.73                             \\ \bottomrule
\end{tabular} 
\caption{Benchmarking PTSM. \textbf{Gold} denotes a model fine-tuned with downstream, original Train data. \textbf{P$n$} indicates that the model is trained on Para$n$ training set. \textbf{P$n$+G} indicates that the model is trained on the combination of Para$n$ and original gold training set. \textbf{Bold} denotes the best result for each task under each group of settings. \textbf{\underline{Underscore}} indicates that the model outperforms the corresponding baseline that is fine-tuned on gold Train set. }~\label{tab:para_res} 
\end{table*}

%% file: tables/no_emoji.tex
\begin{table}[h]
\centering
\tiny
\begin{tabular}{@{}lcc|cc@{}}
\toprule
\multicolumn{1}{c}{\multirow{2}{*}{\textbf{Task}}} & \multicolumn{2}{c}{\textbf{RoBERTa}}                                          & \multicolumn{2}{c}{\textbf{BERTweet}}                     \\\cmidrule(l){2-3}  \cmidrule(l){4-5}  
\multicolumn{1}{c}{}                               & \multicolumn{1}{l}{\textbf{Original}} & \multicolumn{1}{l|}{\textbf{RM emoji}} & \textbf{Original} & \multicolumn{1}{l}{\textbf{RM emoji}} \\ \midrule
Crisis\textsubscript{Oltea}       & {95.96}                                 & 95.88                                 & 95.58             & {95.70}                                 \\
Emo\textsubscript{Moham}          & {77.61}                                 & 77.37                                 & {80.37}             & 79.77                                 \\
Hate\textsubscript{Bas}           & {49.74}                                 & 49.19                                 & {56.51}             & 55.03                                 \\
Hate\textsubscript{Waseem}        & {56.84}                                 & 56.37                                 & {57.07}             & 56.93                                 \\
Hate\textsubscript{David}         & 77.03                                 & {77.19}                                 & {77.57}             & 77.46                                 \\
Humor\textsubscript{Potash}       & {54.66}                                 & 54.60                                 & {57.56}             & 54.47                                 \\
Humor\textsubscript{Meaney}       & {92.61}                                 & 92.38                                 & 94.21             & {94.63}                                 \\
Irony\textsubscript{Hee-A}        & {73.13 }                                & 72.59                                 & 76.82             & {76.97}                                 \\
Irony\textsubscript{Hee-B}        & 51.56                                 & {52.49}                                 & {56.84}             & 56.80                                 \\
Offense\textsubscript{-Zamp}      & {80.67}                                 & 79.38                                 & 79.74             & {79.98}                                 \\
Sarc\textsubscript{Riloff}        & {75.28}                                 & 72.35                                 & 76.61             & {79.60}                                 \\
Sarc\textsubscript{Ptacek}        & 95.59                                 & {95.73}                                 & {96.74}             & 96.45                                 \\
Sarc\textsubscript{Rajad}         & {85.64}                                 & 85.35                                 & {86.97}             & 86.93                                 \\
Sarc\textsubscript{Bam}           & {80.01}                                 & 79.02                                 & {82.59}             & 82.38                                 \\
Senti\textsubscript{Rosen}        & 70.78                                 & 70.44                                 & 72.05             & 71.56                                 \\
Senti\textsubscript{Thel}         & 88.99                                 & 89.94                                 & 89.27             & 89.13                                 \\
Stance\textsubscript{Moham}       & 69.28                                 & 69.64                                 & 69.11             & 66.54                                 \\ \cdashline{1-5}
\textbf{Average}                                            & 75.02                                 & 74.70                                 & 76.80             & 76.49                                 \\ \bottomrule
\end{tabular}
\caption{Effect of emojis in Train. {Original} indicates the original gold Train set. {RM emoji} indicates the gold Train set after removing emojis.   }\label{tab:no_emoji}
\end{table}

%% file: tables/para_test.tex
\begin{table}[h]
\tiny
\centering
\begin{tabular}{@{}lc:cccc@{}}
\toprule
\multicolumn{1}{c}{\textbf{Task}}             & \textbf{Gold} & \textbf{P1} & \textbf{P2} & \textbf{P4} & \textbf{P5} \\ \midrule
Crisis\textsubscript{Oltea}  & 89.67         & 91.87       & 92.04       & 92.11       & 91.92       \\
Emo\textsubscript{Moham}     & 63.40         & 63.64       & 65.25       & 65.35       & 64.36       \\
Hate\textsubscript{Bas}      & 60.83         & 54.19       & 56.85       & 55.82       & 55.85       \\
Hate\textsubscript{Waseem}   & 48.90         & 52.62       & 52.75       & 52.49       & 53.14       \\
Hate\textsubscript{David}    & 58.43         & 66.40       & 67.63       & 66.07       & 67.72       \\
Humor\textsubscript{Potash}  & 54.34         & 53.39       & 52.42       & 49.58       & 49.94       \\
Humor\textsubscript{Meaney}  & 82.89         & 83.92       & 83.78       & 83.17       & 82.82       \\
Irony\textsubscript{Hee-A}   & 63.89         & 66.07       & 66.27       & 67.41       & 66.19       \\
Irony\textsubscript{Hee-B}   & 40.97         & 39.09       & 42.43       & 42.02       & 41.74       \\
Offense\textsubscript{-Zamp} & 73.95         & 72.63       & 73.08       & 74.18       & 74.18       \\
Sarc\textsubscript{Riloff}   & 64.37         & 66.94       & 66.17       & 68.38       & 67.88       \\
Sarc\textsubscript{Ptacek}   & 83.11         & 85.80       & 88.00       & 89.20       & 89.54       \\
Sarc\textsubscript{Rajad}    & 73.46         & 74.85       & 74.44       & 75.21       & 74.87       \\
Sarc\textsubscript{Bam}      & 71.03         & 72.38       & 73.17       & 72.91       & 73.52       \\
Senti\textsubscript{Rosen}   & 64.24         & 63.62       & 64.29       & 64.47       & 65.28       \\
Senti\textsubscript{Thel}    & 84.10         & 83.37       & 83.64       & 84.85       & 83.22       \\
Stance\textsubscript{Moham}  & 62.08         & 62.74       & 62.56       & 63.01       & 62.69       \\ \cdashline{1-6}
\textbf{Average}                                       & 67.04         & 67.85       & 68.52       & 68.60       & 68.52       \\ \bottomrule
\end{tabular}
\caption{Testing on Para-Test. \textbf{Gold} denotes a model fine-tuned with downstream, original Train data. \textbf{P$n$} indicates that the model is trained on Para$n$ training set. }\label{tab:para_test}
\end{table}

%% file: conclusion.tex
\section{Conclusion and Limitations}
Motivated by the issue of decay of social media data, we proposed simple paraphrasing as employed in our PTSM data as a way to avail training datasets for learning social meaning. We fine-tune a T5 model on diverse paraphrasing dataset and utilize the trained model to generate paraphrases of original social meaning Train sets. Through experimental results, we show that we can substitute the actual tweets with their paraphrases while incurring a marginal performance loss. Due to the closed vocabulary of T5, our paraphrasing model cannot generate emojis (even though these can be useful for social meaning detection). We can rectify this by simple post-processing in future work. Another possible limitation of our work is that we use a simple tri-gram similarity method to measure similarity between an actual tweet and its paraphrases. A more sophisticated method may enhance our resulting paraphrase data and hence possibly improve downstream model performance; we will explore this in the future.

\section*{Ethical Considerations}~\label{sec:ethic}
PTSM is collected from publicly available sources and aims at availing resources for training NLP models without need for original user data, which could be a step forward toward protecting privacy. Following Twitter policy, all the data we used for model training are anonymous. We also notice the annotation bias existing in the original datasets (e.g., Hate\textsubscript{Waseem}). 

\section*{Acknowledgements}\label{sec:acknow}
We gratefully acknowledge support from the Natural Sciences and Engineering Research Council of Canada (NSERC; RGPIN-2018-04267), the Social Sciences and Humanities Research Council of Canada (SSHRC; 435-2018-0576; 895-2021-1008), Compute Canada, and UBC ARC-Sockeye.\footnote{\href{https://arc.ubc.ca/ubc-arc-sockeye}{https://arc.ubc.ca/ubc-arc-sockeye}} 